\documentclass{article}

\usepackage[preprint]{neurips_2026}

\usepackage[utf8]{inputenc} 
\usepackage[T1]{fontenc}    
\usepackage{hyperref}       
\usepackage{url}            
\usepackage{booktabs}       
\usepackage{amsfonts}       
\usepackage{nicefrac}       
\usepackage{microtype}      
\usepackage{xcolor}         
\usepackage{graphicx}
\usepackage{float}  
\usepackage{amsmath}
\usepackage{tcolorbox}
\usepackage{wrapfig}
\usepackage{xcolor}
\usepackage{booktabs}
\usepackage{makecell}
\usepackage{algorithm}
\usepackage{algpseudocode}
\usepackage{pifont}
\usepackage{wrapfig}
\usepackage{booktabs}
\usepackage{makecell}
\usepackage{pifont}
\usepackage{graphicx}
\usepackage{needspace}
\usepackage{multirow}
\usepackage{array}
\newcolumntype{P}[1]{>{\raggedright\arraybackslash}p{#1}}

\newcommand{\cmark}{\ding{51}}
\newcommand{\xmark}{\ding{55}}

\def \until{{\mathbf{U}}}

\title{ReasonSTL: Bridging Natural Language and Signal Temporal Logic via Tool-Augmented Process-Rewarded Learning}

\author{
Bowen Ye$^{1,2}$ \quad
Zhijian Li$^{2}$ \quad
Junyue Huang$^{1}$ \quad
Junkai Ma$^{2}$ \quad
Xiang Yin$^{1,*}$ \\[1mm]
$^{1}$Shanghai Jiao Tong University \\
$^{2}$Alibaba Group, Hangzhou, China \\[1mm]
$^{*}$Corresponding author.
}

\begin{document}

\maketitle

\begin{abstract}
Signal Temporal Logic (STL) is an expressive formal language for specifying spatio-temporal requirements over real-valued, real-time signals. It has been widely used for the verification and synthesis of autonomous systems and cyber-physical systems. In practice, however, users often express their requirements in natural language rather than in structured STL formulas, making natural-language-to-STL translation a critical yet challenging task. Manual specification requires temporal-logic expertise and cannot scale, while prompting commercial LLM APIs incurs substantial token costs and may expose sensitive system requirements to third-party services, raising privacy concerns for industrial deployment. To address these challenges, we present \textsc{ReasonSTL}, a tool-augmented framework that adapts local open-source language models for natural-language-to-STL generation. \textsc{ReasonSTL} decomposes the translation process into explicit reasoning, deterministic tool calls, and structured formula construction. We further introduce process-rewarded training to supervise both tool-use trajectories and final formulas, together with \textsc{STL-Bench}, a bilingual, computation-aware benchmark grounded in real-world signals. Experiments show that a 4B model trained with \textsc{ReasonSTL} achieves state-of-the-art performance in both automatic metrics and human evaluations, demonstrating that \textsc{ReasonSTL} provides a transparent, low-cost, and privacy-preserving alternative for formal specification drafting.
\end{abstract}

\begin{figure}[H]
    \centering
    \vspace{-1em}\includegraphics[width=1\linewidth]{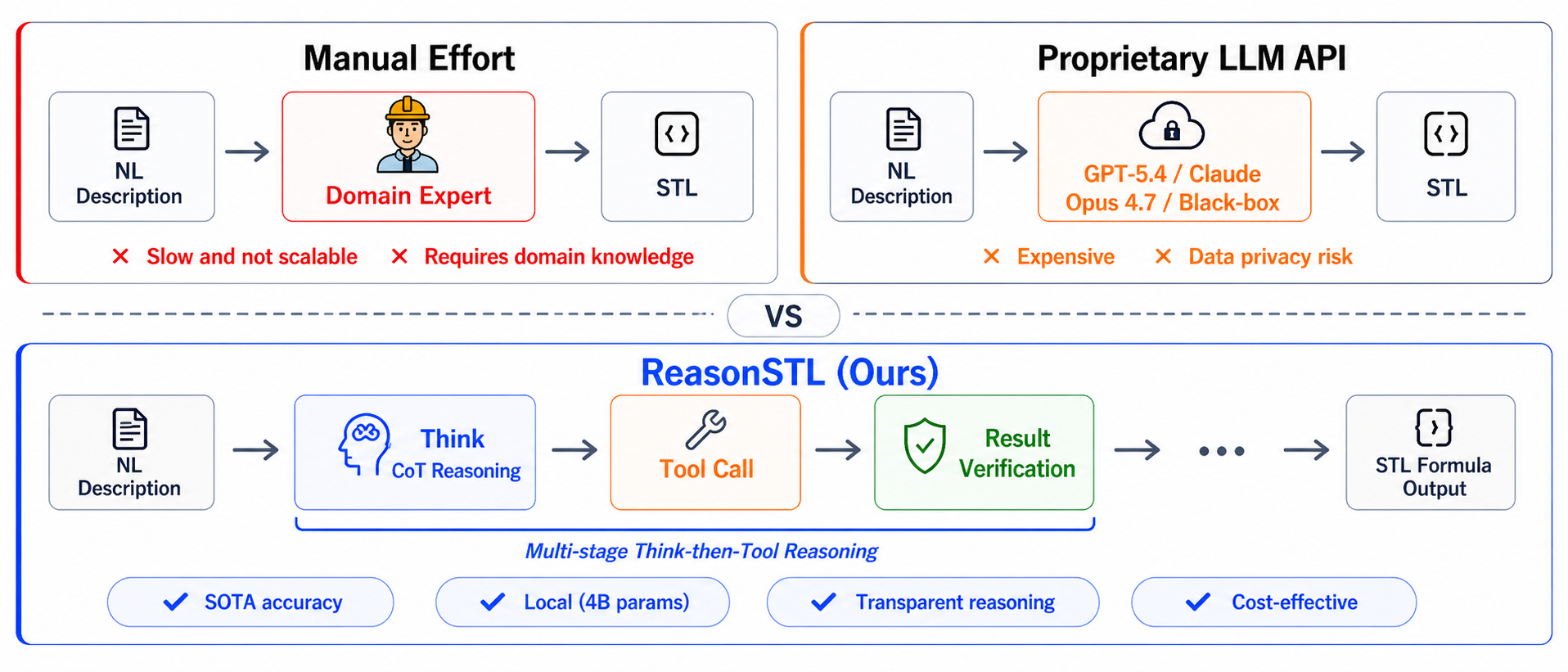}
    \caption{
    Comparison of three Approaches for translating NL descriptions into STL specifications.
    }
    \label{fig:shown}
\end{figure}




















\section{Introduction}
\label{sec:intro}

Formal methods provide a rigorous mathematical foundation for specifying, verifying, monitoring, and synthesizing complex systems~\cite{yin2024formal,belta2019formal}. By formally specifying system requirements, one can reason about safety, performance, and operational correctness with mathematical guarantees. Among formal specification languages, \emph{Signal Temporal Logic} (STL) is particularly suitable for cyber-physical systems (CPS) and autonomous systems, as it expresses temporal properties over continuous, real-valued signals such as distance, velocity, temperature, voltage, pressure, and control inputs~\cite{maler2004monitoring,bartocci2018specification}. STL has been widely applied in monitoring, verification, planning, learning, and control across domains including autonomous driving~\cite{arechiga2019specifying}, unmanned aerial vehicles~\cite{silano2021power}, robotics~\cite{charitidou2024distributed,lindemann2018control}, electronic systems~\cite{kong2016temporal}, and collaborative robots~\cite{yu2024trust}. It also serves as objectives or constraints in optimization, end-to-end learning, zero-shot task allocation, and temporal-logic-guided reinforcement learning~\cite{raman2014model,ye2025bridging,liu2026zeroshot,saxena2023funnel,meng2025tgpo}.

In practice, many CPS tasks are described in natural language by engineers or domain experts, using high-level requirements that encode safety constraints, performance goals, and operational assumptions. Translating these specifications into temporal logic is therefore crucial, as it enables formal verification, automated planning, and synthesis while bridging the gap between informal descriptions and mathematically precise system behavior. Most existing NL-to-TL methods focus on Linear Temporal Logic (LTL) or Boolean specifications, which are suitable for symbolic discrete propositions but do not directly support the continuous, signal-based reasoning required for STL~\cite{brunello2019synthesis,gopalan2018sequence,liu2023lang2ltl,chen2023nl2tl}. This gap motivates the development of reliable NL-to-STL generation capable of handling numerical values, continuous signals, temporal intervals, and domain-specific predicates inherent in CPS requirements.

NL-to-STL translation is technically challenging. STL formulas are executable formal objects, where small errors in operator structure, predicate grounding, temporal intervals, or numerical thresholds can lead to incorrect semantics. Realistic CPS requirements often include colloquial temporal expressions, physical units, arithmetic constraints, event-triggered conditions, and domain-specific signals. For instance, phrases like ``within half an hour'', ``10,000 ft'', or ``200 knots'' must be normalized into precise temporal intervals or numerical predicates, while lexical cues such as ``rise'', ``drop'', or ``surge'' may not correspond directly to formal events. Effective NL-to-STL generation therefore requires semantic grounding, computation-aware reasoning, and structured formula construction beyond mere syntactic validity.

Translating natural-language requirements into STL formulas has traditionally relied on manual specification by engineers or domain experts. While this approach is transparent and allows precise control over the resulting formulas, it is labor-intensive and does not scale to large or complex  tasks. To reduce this burden, recent works have laid important foundations for automated NL-to-STL generation. For example, DeepSTL casts the task as a neural machine translation problem and trains the network from scratch~\cite{he2022deepstl}; however, this approach is constrained by its from-scratch training paradigm, which restricts its generalization across diverse specifications and deployment scenarios. KGST introduces a generate-then-refine pipeline in which a fine-tuned small model first produces candidate formulae, an external knowledge base retrieves relevant examples, and a proprietary LLM API ultimately synthesizes the final STL output~\cite{fang2025enhancing}. Nevertheless, the concurrent involvement of fine-tuned small models and proprietary LLM APIs inevitably entails substantial computational overhead and raises non-trivial privacy concerns, as sensitive data must be transmitted to external third-party services. RESTL further incorporates reinforcement learning with multi-aspect rewards, curriculum learning, and PPO-based optimization~\cite{fang2026restl}. Nevertheless, despite operating entirely on local models, it directly generates STL formulae without any intermediate reasoning process, which may compromise the interpretability and correctness of the outputs. As illustrated in Figure~\ref{fig:shown}, these limitations collectively underscore the need for a principled framework that is fully local, annotation-free, scalable, and capable of producing verifiable intermediate reasoning traces throughout the STL synthesis process.

To address these challenges, we present \textsc{ReasonSTL}, a local tool-augmented framework for computation-aware NL-to-STL generation. Instead of single-step sequence prediction, \textsc{ReasonSTL} decomposes specification drafting into explicit reasoning, deterministic computation, and structured STL construction. The model invokes tools for temporal normalization, unit conversion, arithmetic evaluation, and time-difference computation, then assembles a structured STL JSON tree from verified intermediate results. This decouples linguistic and structural reasoning from exact numerical computation, making the generation process inspectable and amenable to validation. Outcome-bounded process supervision provides feedback on both intermediate tool use and final STL correctness, encouraging the model to learn when deterministic computation is needed and how to integrate results.

We also introduce \textsc{STL-Bench}, a bilingual benchmark for structured and computation-aware NL-to-STL evaluation. It covers domain-grounded CPS signals, physical units, arithmetic constraints, nested temporal structures, structured STL trees, and tool-use annotations, constructed through expert-guided templates, model-assisted requirement realization, rule-based validation, embedding-based semantic pruning, and human auditing. Experiments demonstrate that \textsc{ReasonSTL} achieves state-of-the-art performance on both automatic metrics and human verification, providing a local, verifiable alternative to black-box API-based STL generation.

The main contributions are summarized as follows:
\begin{itemize}
    \item \textbf{Local tool-augmented NL-to-STL generation.}
    We propose \textsc{ReasonSTL}, a local open-source framework that converts natural-language requirements into STL specifications through multi-step reasoning, deterministic tool invocation, and structured formula construction. It explicitly handles temporal normalization, unit conversion, arithmetic evaluation, predicate grounding, and event-operator selection.

    \item \textbf{Outcome-bounded process supervision.}
    We introduce a process-aware training strategy that supervises both intermediate tool-use trajectories and final STL construction. By bounding the maximum intermediate reward with final-formula correctness, it reduces the risk of reinforcing plausible but semantically invalid reasoning traces.

    \item \textbf{Computation-aware benchmark and evaluation.}
    We construct \textsc{STL-Bench}, a bilingual benchmark with domain-grounded CPS signals, physical units, arithmetic constraints, nested temporal structures, and tool-use annotations. Experiments on \textsc{STL-Bench} and existing NL-to-STL datasets show that \textsc{ReasonSTL} achieves state-of-the-art performance, providing a locally deployable alternative to black-box API-based specification generation.
\end{itemize}

\section{Preliminaries}
\label{sec:pre}
\subsection{Signal Temporal Logic}

Let $\mathcal{S}\subseteq\mathbb{R}^d$ denote the signal space, where each dimension corresponds to a real-valued CPS signal variable, and let
$\mathbf{s}_{0:T}=(s_0,\ldots,s_T)\in\mathcal{S}^{T+1}$ denote a finite discrete-time trace. Signal Temporal Logic (STL) provides a formal language for specifying temporal properties over such traces. We consider the following grammar:
\begin{equation}
    \phi ::= \top \mid \pi^\mu \mid \neg\phi
    \mid \phi_1 \wedge \phi_2
    \mid \phi_1 \until_{[a,b]} \phi_2 ,
\end{equation}
where $\pi^\mu$ is an atomic predicate induced by a function $\mu:\mathcal{S}\to\mathbb{R}$ and is satisfied at time $k$ iff $\mu(s_k)\ge 0$. The temporal interval bounds satisfy $a,b\in\mathbb{N}$ and $0\le a\le b$.

The bounded until operator is interpreted as
\begin{equation}
(\mathbf{s},k)\models \phi_1\until_{[a,b]}\phi_2
\end{equation}
iff there exists a time index $k'\in[k+a,k+b]$ such that $(\mathbf{s},k')\models\phi_2$ and $(\mathbf{s},k'')\models\phi_1$ for all $k''\in[k,k')$. The standard derived operators ``eventually'' and ``always'' are defined as
\begin{equation}
    \mathbf{F}_{[a,b]}\phi := \top\until_{[a,b]}\phi,
    \qquad
    \mathbf{G}_{[a,b]}\phi := \neg\mathbf{F}_{[a,b]}\neg\phi .
\end{equation}
We write $\mathbf{s}\models\phi$ as shorthand for $(\mathbf{s},0)\models\phi$. And \textsc{STL-Bench} also includes common CPS-oriented syntactic extensions, such as threshold-crossing predicates \emph{rise} and \emph{fall}, and the past-time operator $\mathbf{H}_{[a,b]}$ (\emph{historically}). We treat them as derived constructs with fixed semantics, with formal definitions provided in Appendix~\ref{app:stl_details}.

\subsection{Natural-Language-to-STL Translation}
\label{sec:nl2stl_task}

Given a natural-language requirement $q$, the goal is to produce an STL formula $\phi$ that preserves the intended temporal, logical, and numerical semantics of $q$. Let
\begin{equation}
    \mathcal{D}=\{(q_i,\phi_i^\star)\}_{i=1}^{N}
\end{equation}
be a dataset of requirements and reference formulas. A model $\pi_\theta$ generates an output
\begin{equation}
    \hat{y}_i \sim \pi_\theta(\cdot\mid q_i),
\end{equation}
which is parsed into a predicted formula $\hat{\phi}_i$ when it satisfies the target syntax.

NL-to-STL translation differs from ordinary text generation because small structural or numerical errors can change the meaning of the specification. A valid prediction must identify signal variables, comparison operators, thresholds, Boolean structure, temporal operators, and time intervals. Many requirements also involve implicit computations, such as duration normalization, unit conversion, or arithmetic threshold calculation, before the final formula can be constructed.

\paragraph{Structured representation.}
For training and evaluation, we serialize STL formulas as JSON trees. Internal nodes represent temporal or Boolean operators, and leaves represent atomic predicates. For example,
\[
\mathbf{F}_{[0,1800]}
\big((\mathrm{altitude}>3048.0)\wedge(\mathrm{speed}\ge102.89)\big)
\]
is represented as a tree rooted at \texttt{Finally} with interval $[0,1800]$, whose child is an \texttt{and} node over two predicates.

This representation removes ambiguity from parentheses and operator precedence, supports schema validation, and enables recursive comparison of formula structure. Importantly, the JSON representation is an evaluation interface rather than an additional semantic language: each valid JSON tree corresponds to an STL formula. A concrete schema example is given in Appendix~\ref{app:json_schema}.

\paragraph{Evaluation metrics.}
We report two automatic metrics. 

\emph{Format Accuracy} measures whether the generated STL formula matches the reference at the structural level after abstracting away fine-grained predicate and numerical details. Specifically, we first parse the model output into a structured STL tree and then apply a masking function $\mathcal{M}(\cdot)$ that replaces predicate contents, temporal bounds, and numerical thresholds with abstract placeholders. Format Accuracy is defined as
\begin{equation}
    \mathrm{FormatAcc}
    =
    \frac{1}{N}
    \sum_{i=1}^{N}
    \mathbb{I}
    \left[
        \hat{y}_i\in\mathcal{Y}_{\mathrm{valid}}
        \wedge
        \mathrm{Match}_{\mathrm{STL}}
        \bigl(
            \mathcal{M}(\hat{\phi}_i),
            \mathcal{M}(\phi_i^\star)
        \bigr)
    \right],
\end{equation}
where $\mathcal{Y}_{\mathrm{valid}}$ denotes the set of outputs that can be parsed into the required structured STL representation. Thus, Format Accuracy evaluates whether the model produces the correct formula skeleton, including the temporal and logical operator structure, while ignoring exact predicate grounding and numerical values.

\emph{Formula Accuracy} is a stricter metric that requires the generated formula to fully match the reference formula:
\begin{equation}
    \mathrm{FormulaAcc}
    =
    \frac{1}{N}
    \sum_{i=1}^{N}
    \mathbb{I}
    \left[
        \hat{y}_i\in\mathcal{Y}_{\mathrm{valid}}
        \wedge
        \mathrm{Match}_{\mathrm{STL}}(\hat{\phi}_i,\phi_i^\star)
    \right].
\end{equation}
Here, the matcher recursively compares STL formula trees, treats commutative Boolean operators as order-invariant, and applies a fixed numerical tolerance to temporal bounds and numerical thresholds. Each sample receives a binary score: a correct match is assigned $1$, and an incorrect match is assigned $0$. Since automatic tree matching may not capture all semantic equivalences between STL formulas, we complement it with human verification in Section~\ref{sec:exp}.

\section{\textsc{STL-Bench}: A Computation-Aware NL-to-STL Benchmark}
\label{sec:bench}

\paragraph{Benchmark overview.}
Existing NL-to-STL datasets have enabled early progress in data-driven specification generation, with DeepSTL providing a grammar-guided English--STL corpus~\cite{he2022deepstl} and STL-DivEn improving linguistic diversity through LLM-assisted augmentation and validation~\cite{fang2025enhancing}. However, they provide limited coverage of computation-sensitive CPS requirements involving domain-grounded predicates, temporal normalization, physical units, arithmetic thresholds, event-triggered semantics, and multilingual descriptions. To address this gap, \textsc{STL-Bench} is designed around three principles: \emph{domain grounding}, where predicates are defined over physically meaningful CPS signals rather than abstract symbols; \emph{computation awareness}, where time expressions, units, and arithmetic thresholds are explicitly normalized before formula construction; and \emph{structured evaluation}, where STL specifications are represented as JSON trees to support schema validation, recursive structural matching, and diagnostic analysis of intermediate tool use.

Tables~\ref{tab:benchmark_summary} and~\ref{tab:dataset_comparison} summarize the benchmark statistics and its comparison with representative NL-to-STL datasets. \textsc{STL-Bench} contains 28,880 bilingual NL--STL pairs across six engineering domains and 33 CPS scenarios, with 73\% of samples requiring at least one intermediate computation. Compared with prior datasets, \textsc{STL-Bench} evaluates not only final-formula correctness, but also predicate grounding, temporal normalization, unit conversion, arithmetic reasoning, event-operator grounding, bilingual understanding, and structured formula construction.

\begin{table}[t]
\centering
\small
\caption{Summary of \textsc{STL-Bench} construction and evaluation splits.}
\label{tab:benchmark_summary}
\resizebox{\linewidth}{!}{
\begin{tabular}{lll}
\toprule
\textbf{Property} & \textbf{Value} & \textbf{Description} \\
\midrule
Total samples & 28,880 & Bilingual NL--STL pairs \\
Languages & 2 & English and Chinese \\
Domains & 6 & Industrial, automotive, aerospace, robotics, environmental, electrical \\
Scenarios & 33 & Domain-specific CPS settings \\
Signals & 41 & Domain-grounded CPS variables \\
Complexity & 1--6 & Formula and tool-use complexity \\
Tool-use samples & 73\% & Require intermediate computation \\
\texttt{parse\_duration} & 7,337 & Temporal normalization \\
\texttt{convert\_unit} & 5,161 & Unit conversion \\
\texttt{eval\_math\_expr} & 4,527 & Arithmetic evaluation \\
Deduplication & Symbolic + embedding & Structural and semantic filtering \\
Standard split & 8:1:1 & Train / validation / test \\
Held-out split & Scenario-level & Unseen scenarios \\
Manual test set & $\sim$200 & Human-written, template-free \\
Human auditing & Stratified & Language, domain, complexity, and tool type \\
\bottomrule
\end{tabular}
}
\end{table}

\begin{table}[t]
\centering
\small
\caption{Comparison with representative NL-to-STL datasets.}
\label{tab:dataset_comparison}
\resizebox{\linewidth}{!}{
\begin{tabular}{lccc}
\toprule
\textbf{Dataset Property}
& \textbf{DeepSTL~\cite{he2022deepstl}}
& \textbf{STL-DivEn~\cite{fang2025enhancing}}
& \textbf{\textsc{STL-Bench (Ours)}} \\
\midrule
Language & English & English & English + Chinese \\
Construction & Grammar-based & Seed + LLM + validation & Scenario + template + LLM + validation \\
Domain grounding & No & Partial & 6 domains / 33 scenarios \\
Signal variables & Symbolic & Mostly symbolic & CPS-semantic variables \\
Output format & STL string & STL string & Structured JSON tree \\
Intermediate computation & No & No & Explicitly annotated \\
Tool-use trajectory & No & No & 73\% samples \\
Complexity annotation & No & No & 1--6 levels \\
Semantic deduplication & No & Partial & Symbolic + embedding-based pruning \\
Split protocol & Not emphasized & Not emphasized & Standard + held-out + manual \\
Human validation & No & Yes & Stratified audit \\
\bottomrule
\end{tabular}
}
\end{table}

\paragraph{Construction and validation.}
\textsc{STL-Bench} is constructed through a template-anchored and verification-driven pipeline. We first define CPS domains, scenario categories, semantic signal variables, and parameterized STL templates. Each template specifies admissible operators, signal variables, time intervals, thresholds, and optional unit systems. A strong language model is then used for controlled requirement realization: it generates engineering-style natural-language descriptions conditioned on the fixed scenario, signal semantics, numerical parameters, and target formula structure, while the reference STL formula is determined by the template and parameter assignment before text generation. This design separates formal-label construction from linguistic realization and reduces semantic drift.

During annotation, we distinguish threshold-state requirements from edge-triggered event requirements. Expressions such as ``rises above'', ``drops below'', or ``surges past'' are not automatically mapped to \emph{rise} or \emph{fall}; they are annotated as edge events only when the requirement explicitly specifies a transition from violation to satisfaction, or vice versa. For computation-sensitive samples, we additionally annotate tool-use trajectories, including the tool type, input arguments, expected outputs, and the final STL fields where computed values are used.

All samples are processed by rule-based validators before inclusion. The validators check JSON syntax, STL-tree well-formedness, operator arity, temporal-interval validity, predicate format, signal-variable validity, unit-conversion consistency, and alignment between tool outputs and final formula values. Invalid samples are removed or repaired only when the correction is deterministic. We further apply symbolic duplicate filtering over STL structures and predicate assignments, followed by embedding-based semantic near-duplicate pruning using Qwen3-Embedding-4B. Finally, a stratified subset is manually audited across languages, domains, complexity levels, and tool-use types.

\paragraph{Split protocol.}
After deduplication, we construct train, validation, and test sets using fixed split. In addition to this standard partition, we build a scenario-aware split in which a subset of scenarios from each high-level engineering domain is held out from training and used only for validation or testing. This protocol reduces direct reuse of identical domain--template combinations while preserving broad coverage for controlled model comparison.

To further assess generalization beyond the template-anchored generation process, we construct an additional manually written test set of approximately 200 requirements. These samples are authored by human annotators in both English and Chinese without using the data-generation templates, paired with manually specified structured STL formulas, and checked by the same validation pipeline. We use this set as an extra held-out evaluation and report its results together with the standard and scenario-aware splits in Section~\ref{sec:exp}.

\begin{figure}[t]
    \centering
    \includegraphics[width=1\linewidth, trim=0 0 0 0, clip]{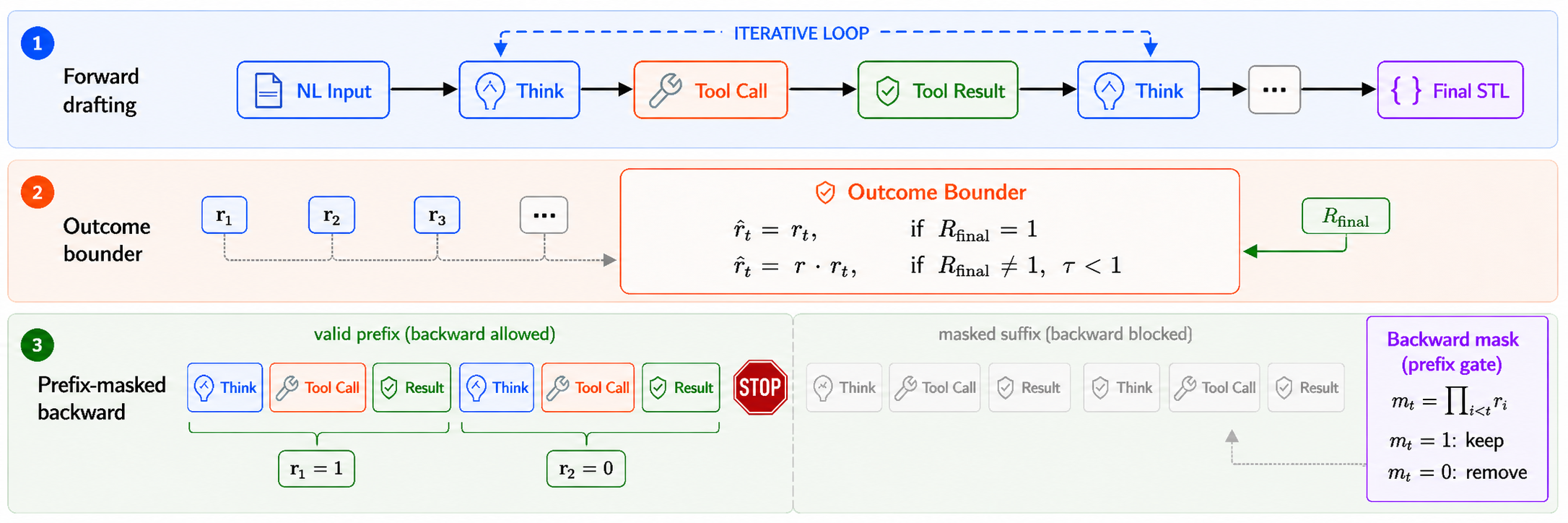}
    \caption{
    Overview of \textsc{ReasonSTL}. The framework converts natural-language requirements into structured STL specifications through interleaved reasoning and deterministic tool execution, and is optimized with outcome-bounded process rewards to supervise both intermediate tool use and final formula construction.
    }
    \label{fig:framework}
\end{figure}

\section{Method}
\label{sec:method}

This section presents \textsc{ReasonSTL}, a local tool-augmented framework for computation-aware NL-to-STL generation. As shown in Figure~\ref{fig:framework}, \textsc{ReasonSTL} formulates specification generation as a structured decision process: temporal bounds, physical units, arithmetic thresholds, and event predicates are explicitly resolved through intermediate reasoning and deterministic tool calls before being assembled into a structured STL specification. The framework is trained with outcome-bounded process rewards, where intermediate rewards are capped by final-formula correctness and assigned only to valid reasoning prefixes. This encourages the model to construct correct STL formulas through faithful step-by-step computation rather than rewarding invalid intermediate trajectories.

\subsection{Tool-Augmented STL Generation}
\label{sec:method_framework}

Given a natural-language requirement $q$, the model generates a structured STL formula $\hat{\phi}$ in the JSON-tree representation defined in Section~\ref{sec:nl2stl_task}. A rollout is written as
\begin{equation}
    y=(z_1,u_1,o_1,\ldots,z_M),
\end{equation}
where $z_m$ is the model-generated segment at stage $m$, $u_m$ is an optional tool call, and $o_m$ is the deterministic tool output. The final segment contains the predicted STL JSON. \textsc{ReasonSTL} uses a compact typed tool set
\begin{equation}
    \mathcal{T}=\{
    \texttt{parse\_duration},
    \texttt{convert\_unit},
    \texttt{eval\_math\_expr},
    \texttt{calc\_time\_diff}
    \},
\end{equation}
covering temporal normalization, unit conversion, arithmetic evaluation, and time-difference computation. This interface assigns exact numerical computation to deterministic tools, while leaving semantic interpretation, tool selection, and STL composition to the language model.

\subsection{Process-Rewarded Optimization}
\label{sec:method_reward}

Tool-augmented generation makes each rollout decomposable into reasoning segments, tool calls, tool outputs, and final STL construction. This enables fine-grained supervision, but also creates a structured credit-assignment problem: an incorrect formula may arise from invalid tool selection, malformed arguments, inconsistent tool outputs, unjustified event-operator grounding, ill-formed JSON, or an incorrect STL tree. \textsc{ReasonSTL} therefore combines outcome-level STL rewards with outcome-bounded process rewards and prefix-masked backward supervision.

For a requirement $q$, we sample a group of $G$ rollouts from the old policy $\pi_{\theta_{\mathrm{old}}}$. Each rollout $y_i$ is parsed into $K_i$ intermediate stages and a final structured STL prediction $\hat{\phi}_i$. A deterministic validator assigns a binary local correctness score
\begin{equation}
    c_{i,k}\in\{0,1\}, \qquad k=1,\ldots,K_i,
\end{equation}
to each intermediate stage, checking tool validity, argument format, executability, output consistency, and semantic correctness. Event operators such as \emph{rise} and \emph{fall} are rewarded only when the requirement explicitly specifies a transition, rather than being inferred from surface lexical cues.

The final STL prediction is evaluated by recursive tree matching. Let
\begin{equation}
    S_i^{\mathrm{tree}}
    =
    \mathrm{TreeMatch}(\hat{\phi}_i,\phi^\star)
    \in[0,1],
\end{equation}
where the matcher $\mathrm{TreeMatch}$ compares operators, intervals, predicates, and subformulas, with order-invariant matching for commutative Boolean operators and numerical tolerance for time bounds and thresholds. If the output cannot be parsed into a valid JSON tree, we set $R_i^{\mathrm{fmt}}=0$ and $S_i^{\mathrm{tree}}=0$. To preserve a strict preference for exact correctness, non-perfect tree matches are capped:
\begin{equation}
R_i^{\mathrm{cnt}}
=
\begin{cases}
1, & S_i^{\mathrm{tree}}=1,\\
\kappa S_i^{\mathrm{tree}}, & S_i^{\mathrm{tree}}<1,
\end{cases}
\qquad \kappa<1 .
\label{eq:capped_tree_reward}
\end{equation}
The outcome reward and final exact-correctness indicator are defined as
\begin{equation}
    R_i^{\mathrm{out}} = R_i^{\mathrm{fmt}} R_i^{\mathrm{cnt}},
    \qquad
    C_i^{\mathrm{final}}
    =
    \mathbb{I}\!\left[
    R_i^{\mathrm{fmt}}=1
    \wedge
    S_i^{\mathrm{tree}}=1
    \right],
    \label{eq:outcome_reward}
\end{equation}
where $R_i^{\mathrm{fmt}}\in\{0,1\}$ indicates JSON/schema validity.

We next bound process rewards by final-formula correctness. Even if an intermediate step is locally plausible, it should not receive full credit when the final STL formula is incorrect. Meanwhile, because later stages depend on earlier computations, process supervision is applied only to the valid prefix of a rollout. The effective process reward is
\begin{equation}
    R_{i,k}^{\mathrm{proc}}
    =
    \underbrace{\prod_{\ell<k} c_{i,\ell}}_{\text{prefix mask}}
    \cdot
    \underbrace{
    \bigl(C_i^{\mathrm{final}}+\tau(1-C_i^{\mathrm{final}})\bigr)
    }_{\text{outcome bounder}}
    \cdot c_{i,k},
    \qquad 0<\tau<1 ,
    \label{eq:effective_process_reward}
\end{equation}
where the empty product is defined as $1$. Thus, exact final correctness allows valid intermediate stages to receive full process reward, while incorrect final formulas cap the maximum attainable process reward. Once an intermediate stage fails, all subsequent dependent stages are masked from process-level backward updates.

Finally, we compute group-relative advantages separately for final outcomes and intermediate stages:
\begin{equation}
    A_i^{\mathrm{out}}
    =
    \frac{
        R_i^{\mathrm{out}}-\mathrm{mean}_{j=1}^{G}R_j^{\mathrm{out}}
    }{
        \mathrm{std}_{j=1}^{G}R_j^{\mathrm{out}}+\epsilon
    },
    \qquad
    A_{i,k}^{\mathrm{proc}}
    =
    \frac{
        R_{i,k}^{\mathrm{proc}}-\mathrm{mean}_{j\in\mathcal{G}_k}R_{j,k}^{\mathrm{proc}}
    }{
        \mathrm{std}_{j\in\mathcal{G}_k}R_{j,k}^{\mathrm{proc}}+\epsilon
    },
    \label{eq:separate_advantages}
\end{equation}
where $\mathcal{G}_k$ contains rollouts whose parsed trajectories include the corresponding aligned stage. Final STL predictions are therefore normalized against final predictions, while intermediate tool-use stages are normalized against corresponding intermediate stages. Each token inherits the advantage of its parsed segment:
\begin{equation}
    A_{i,t}
    =
    \begin{cases}
    A_{i,k}^{\mathrm{proc}}, & y_{i,t}\ \text{belongs to intermediate stage } k,\\
    A_i^{\mathrm{out}}, & y_{i,t}\ \text{belongs to final STL construction}.
    \end{cases}
    \label{eq:token_advantage}
\end{equation}

We optimize the group-relative policy objective
\begin{equation}
    \mathcal{J}_{\mathrm{PR}}(\theta)
    =
    \mathbb{E}
    \left[
    \frac{1}{G}
    \sum_{i=1}^{G}
    \frac{1}{T_i}
    \sum_{t=1}^{T_i}
    M_{i,t}
    \rho_{i,t}(\theta)
    A_{i,t}
    \right],
    \qquad
    \rho_{i,t}(\theta)
    =
    \frac{
        \pi_\theta(y_{i,t}\mid q,y_{i,<t})
    }{
        \pi_{\theta_{\mathrm{old}}}(y_{i,t}\mid q,y_{i,<t})
    },
    \label{eq:process_objective}
\end{equation}
where $M_{i,t}$ masks invalid, truncated, prefix-blocked, and deterministic tool-output tokens. We use group-normalized advantages and gradient clipping for stability, without PPO-style likelihood-ratio clipping. Additional implementation details are provided in Appendix~\ref{app:method_details}.

\section{Experiments}
\label{sec:exp}

We evaluate \textsc{ReasonSTL} on two benchmarks. DeepSTL~\cite{he2022deepstl} serves as a standard NL-to-STL benchmark for comparison with prior task-specific systems. \textsc{STL-Bench} evaluates a more computation-aware setting involving bilingual requirements, structured JSON formulas, domain-grounded CPS predicates, and explicit intermediate computations.

\subsection{Experimental Setup}
\label{sec:exp_setup}

We report \emph{Formula Accuracy} and \emph{Format Accuracy}, as defined in Section~\ref{sec:nl2stl_task}. Formula Accuracy measures whether the parsed and canonicalized prediction matches the reference specification, whereas Format Accuracy evaluates whether the output satisfies the required STL representation. Black-box API baselines are evaluated with carefully designed prompts, few-shot demonstrations, and explicit output-format instructions. Unless otherwise specified, API models are decoded with temperature $0.1$; Kimi-K2.6 is evaluated with temperature $1.0$ because lower-temperature decoding is not supported by the provider. Local models are evaluated deterministically with temperature $0.0$. We provide additional standard deviations, ablations, and implementation details in Appendix~\ref{app:additional_experiments}.

Local training is conducted on $8\times$H20 GPUs, and all local evaluations are performed on a single H20 GPU. On \textsc{STL-Bench}, \textsc{ReasonSTL} first undergoes an SFT cold start to initialize the structured JSON output format and the starts reinforcement learning. \textsc{ReasonSTL-Direct} is trained for 3K steps and takes approximately 20 hours, whereas the full \textsc{ReasonSTL} model is trained for 5K RL steps and takes approximately 5 days. Detailed comparisons between SFT-only, RL-only, and SFT-initialized RL variants are provided in Appendix~\ref{app:sft_rl_ablation}.

\subsection{Results on DeepSTL}
\label{sec:exp_deepstl}

\begin{wraptable}[12]{r}{0.50\linewidth}
\vspace{-1.2em}
\centering
\small
\caption{Performance on DeepSTL.}
\label{tab:deepstl_results}
\resizebox{\linewidth}{!}{
\begin{tabular}{lcc}
\toprule
\textbf{Method} & \textbf{Formula Acc.} & \textbf{Format Acc.} \\
\midrule
\textbf{ReasonSTL-Direct} & \textbf{0.8143} & \textbf{0.8257} \\
Claude-Opus-4.7 & 0.6603 & 0.6691 \\
RESTL~\cite{fang2026restl} & 0.5985 & 0.6327 \\
DeepSeek-V4 & 0.5685 & 0.5758 \\
GPT-5.4 & 0.5150 & 0.5220 \\
Kimi-K2.6 & 0.5048 & 0.5113 \\
KGST~\cite{fang2025enhancing} & 0.4538 & 0.4939 \\
DeepSTL~\cite{he2022deepstl} & 0.2002 & 0.2916 \\
Qwen3-4B & 0.1131 & 0.1631 \\
\bottomrule
\vspace{-1.8em}
\end{tabular}
}
\end{wraptable}
Table~\ref{tab:deepstl_results} reports results on the DeepSTL benchmark. Since DeepSTL does not provide intermediate-computation annotations or tool-use trajectories, we evaluate \textsc{ReasonSTL-Direct}, a Qwen3-4B-based direct-generation variant without explicit think traces or tool calls, to isolate the effect of task-specific adaptation and structured STL generation. \textsc{ReasonSTL-Direct} achieves the best performance among all compared methods, reaching 81.43 Formula Accuracy and 0.8257 Format Accuracy. It outperforms the strongest black-box API baseline, Claude-Opus-4.7, by more than 15 absolute points on both metrics, and improves over RESTL by 21.58 points in Formula Accuracy and 19.30 points in Format Accuracy. The large gap over the untuned Qwen3-4B backbone indicates that general instruction-following ability alone is insufficient for reliable STL generation. Overall, these results show that task-specific adaptation can move local NL-to-STL generation toward a practically usable regime, where most generated specifications are structurally valid and formula-correct under automatic evaluation.

\subsection{Results on \textsc{STL-Bench}}
\label{sec:exp_stlbench}

Table~\ref{tab:stlbench_results} reports automatic evaluation on \textsc{STL-Bench}. The base Qwen3-4B model performs poorly in both languages, showing that general instruction-following ability alone is insufficient for structured NL-to-STL generation. Task-specific adaptation brings substantial gains: \textsc{ReasonSTL-Direct} improves Formula Accuracy from 0.070 to 0.420 on English and from 0.090 to 0.330 on Chinese. The full \textsc{ReasonSTL} model further improves to 0.510 and 0.470, respectively, indicating that explicit reasoning, deterministic tool use, and process-level supervision are beneficial for computation-sensitive requirements.

\begin{wraptable}[11]{r}{0.56\linewidth}
\vspace{-1.4em}
\centering
\footnotesize
\setlength{\tabcolsep}{4pt}
\renewcommand{\arraystretch}{1.00}
\caption{Automatic evaluation on \textsc{STL-Bench}.}
\label{tab:stlbench_results}
\begin{tabular}{lcc|cc|cc}
\toprule
\multirow{2}{*}{\textbf{Method}}
& \multirow{2}{*}{\textbf{Think}}
& \multirow{2}{*}{\textbf{Tool}}
& \multicolumn{2}{c|}{\textbf{English}}
& \multicolumn{2}{c}{\textbf{Chinese}} \\
\cmidrule(lr){4-5}\cmidrule(lr){6-7}
&
&
& \textbf{Form.}
& \textbf{Fmt.}
& \textbf{Form.}
& \textbf{Fmt.} \\
\midrule
Qwen3-4B
& \xmark & \xmark
& 0.070 & 0.110
& 0.090 & 0.130 \\
ReasonSTL-Direct
& \xmark & \xmark
& 0.420 & 0.500
& 0.330 & 0.410 \\
\textbf{ReasonSTL}
& \cmark & \cmark
& \textbf{0.510} & \textbf{0.560}
& \textbf{0.470} & \textbf{0.490} \\
\midrule
DeepSeek-V4
& -- & --
& 0.310 & 0.330
& 0.270 & 0.350 \\
GPT-5.4
& -- & --
& 0.220 & 0.240
& 0.190 & 0.200 \\
Claude-Opus-4.7
& -- & --
& 0.260 & 0.280
& 0.270 & 0.280 \\
Kimi-K2.6
& -- & --
& 0.170 & 0.200
& 0.190 & 0.200 \\
\bottomrule
\end{tabular}
\end{wraptable}
Under automatic matching, \textsc{ReasonSTL} also outperforms carefully prompted API baselines, suggesting stronger local adaptation to schema-constrained and canonical STL generation. Since automatic matching may undercount semantically correct formulas that differ from the reference through harmless surface variations, predicate naming differences, or equivalent reformulations, we complement it with human semantic verification below.

\begin{wraptable}[6]{r}{0.38\linewidth}
\vspace{-1em}
\centering
\footnotesize
\caption{Human Verification.}
\label{tab:manual_verification}
\setlength{\tabcolsep}{14pt}
\renewcommand{\arraystretch}{1.05}
\begin{tabular}{@{}lcc@{}}
\toprule
\textbf{Method}
& \textbf{EN}
& \textbf{CN} \\
\midrule
\textbf{ReasonSTL}
& \textbf{0.53}
& \textbf{0.51} \\
DeepSeek-V4
& 0.50
& \textbf{0.51} \\
Claude-Opus-4.7
& \textbf{0.53}
& 0.49 \\
\bottomrule
\end{tabular}
\end{wraptable}

To assess whether strict tree matching underestimates semantically valid but non-canonical predictions, we conduct human verification on 100 randomly sampled cases from each language split. A prediction is judged correct if it preserves the intended requirement semantics, even when it differs from the reference through benign surface variations, predicate renaming, or logically equivalent reformulations. As shown in Table~\ref{tab:manual_verification}, \textsc{ReasonSTL} remains competitive with strong proprietary models, matching the best English result and achieving the best or tied-best Chinese result.

\begin{wraptable}[8]{r}{0.45\linewidth}
\vspace{-1.2em}
\centering
\footnotesize
\caption{Results on the manually written test set and inference throughput.}
\label{tab:manual_test_latency}
\setlength{\tabcolsep}{10pt}
\renewcommand{\arraystretch}{1.05}
\begin{tabular}{@{}lcc@{}}
\toprule
\textbf{Method}
& \textbf{Form.}
& \textbf{Throughput (qps)} \\
\midrule
\textbf{ReasonSTL}
& \textbf{0.54}
& 2.3 \\
ReasonSTL-Direct
& 0.42
& \textbf{6.1} \\
DeepSeek-V4
& 0.51
& 0.6 \\
Claude-Opus-4.7
& 0.53
& 2.2 \\
\bottomrule
\end{tabular}
\vspace{-0.8em}
\end{wraptable}

We further evaluate on a manually written, template-free test set to examine whether models generalize beyond the formula templates used during benchmark construction. As shown in Table~\ref{tab:manual_test_latency}, \textsc{ReasonSTL} achieves the highest Formula Accuracy, slightly outperforming strong API baselines while achieving higher throughput than Claude-Opus-4.7. \textsc{ReasonSTL-Direct} achieves the highest throughput, making it attractive when efficiency is prioritized, although its lower accuracy indicates that explicit reasoning and tool execution remain important for more reliable specification drafting.

\subsection{Diagnostic Analysis}
\label{sec:exp_diag}
Beyond aggregate accuracy, we analyze representative failures to reveal method-specific error patterns. Carefully prompted API models often hallucinate operators, mis-handle numerical conversions, or confuse signal grounding, whereas \textsc{ReasonSTL} mainly fails under ambiguous event semantics or non-canonical predicate grounding. Detailed analyses and SFT/RL ablations are provided in Appendix~\ref{app:additional_experiments}.

\section{Conclusion}

This paper presented \textsc{ReasonSTL}, a local tool-augmented framework that transforms natural-language requirements into STL specifications through explicit reasoning, deterministic tool use, and structured formula construction. We further introduced outcome-bounded process supervision to jointly guide intermediate tool-use trajectories and final STL construction, while reducing the risk of reinforcing plausible but semantically invalid reasoning traces. We also introduced \textsc{STL-Bench}, a bilingual computation-aware benchmark covering domain-grounded signals, physical units, arithmetic constraints, nested temporal structures, intermediate computation, and structured STL trees. Experiments show that \textsc{ReasonSTL} achieves state-of-the-art automatic matching performance and remains competitive with strong proprietary models under human verification. These results highlight local tool-augmented generation with process-aware supervision as a practical and privacy-conscious direction for reliable NL-to-STL specification acquisition.

\newpage

\bibliographystyle{abbrv}
\bibliography{reference}


\newpage
\appendix
\section{Related Work}
\label{app:related_work}

\subsection{Natural Language to Temporal Logic}

Natural-language-to-temporal-logic (NL-to-TL) translation has long been studied as a way to bridge informal requirements and formal reasoning. Brunello et al.~\cite{brunello2019synthesis} provide an overview of LTL formula synthesis from natural-language texts, highlighting core challenges such as linguistic ambiguity, proposition grounding, and the need to preserve formal semantics. Early neural approaches formulate the task as sequence-to-sequence translation from natural-language commands to non-Markovian task specifications, typically represented using LTL or related temporal-logic formalisms~\cite{gopalan2018sequence}. These works demonstrate the feasibility of learning mappings from language to temporal logic, but they mainly operate over symbolic propositions and discrete task structures.

More recent studies have explored data-driven and LLM-assisted NL-to-TL generation. NL2TL studies the transformation of natural-language requirements into temporal logic with the aid of large language models~\cite{chen2023nl2tl}. Pan et al.~\cite{pan2023data} investigate data-efficient learning of NL-to-LTL translators for robot task specification, while Yang et al.~\cite{yang2024harnessing} and Mendoza et al.~\cite{mendoza2024translating} further examine LLM-based temporal-logic translation and formal-requirement generation. These studies establish NL-to-TL translation as a broad and increasingly active research direction. However, most existing work focuses on LTL or related Boolean temporal specifications, where atomic propositions are typically symbolic and do not require explicit handling of real-valued signals, physical units, numerical thresholds, or temporal normalization.

\subsection{NL-to-LTL Systems and Interactive Specification Drafting}

Several systems aim to make NL-to-LTL translation more practical for users. Fuggitti et al.~\cite{fuggitti2023nl2ltl} introduce an NL2LTL system for converting natural-language instructions into LTL formulas, emphasizing tool usability and integration. Cosler et al.~\cite{cosler2023nl2spec} study interactive translation from unstructured natural language to temporal-logic specifications, where user feedback is used to refine or disambiguate generated formulas. These works are closely related to human-in-the-loop formal specification drafting, as they lower the barrier between informal requirements and temporal-logic representations.

In robotics, LTL has also been widely used as a structured target language for task planning. Gopalan et al.~\cite{gopalan2018sequence} translate natural-language instructions into non-Markovian task specifications, while Lang2LTL maps robot commands to LTL formulas grounded in planning-relevant propositions~\cite{liu2023lang2ltl}. Such methods show that natural language can be connected to formal task specifications for embodied agents. Nevertheless, their formal targets are primarily LTL formulas over symbolic propositions. In contrast, CPS-oriented STL generation must reason over continuous signals and numerical constraints, making the translation problem more computation-sensitive.

\subsection{Natural Language to Signal Temporal Logic}
Compared with NL-to-LTL, NL-to-STL generation has received relatively limited attention, despite the central role of STL in specifying real-valued and time-constrained CPS requirements. DeepSTL formulates NL-to-STL generation as a neural machine translation problem and introduces an early grammar-guided English--STL benchmark~\cite{he2022deepstl}. This work provides an important starting point for data-driven STL generation, but its requirements are largely grammar-controlled and offer limited coverage of realistic, computation-sensitive CPS specifications.

Beyond direct sequence-to-sequence translation, Mohammadinejad et al. propose an interactive and explainable framework for translating natural-language robot task descriptions into STL formulas~\cite{mohammadinejad2024systematic}. Their method combines semantic parsing, pretrained language models, user-in-the-loop clarifications, and a small number of demonstrations to resolve ambiguities in natural-language commands. While this interaction-driven design improves interpretability and ambiguity resolution, it still relies on additional human feedback or demonstrations, which limits its scalability for fully automated and large-scale STL specification drafting.

KGST proposes a generate-then-refine framework in which a fine-tuned small model produces initial candidates, an external knowledge base retrieves relevant examples, and these are jointly fed as contextual prior information to a proprietary LLM API for final STL formula synthesis~\cite{fang2025enhancing}. By leveraging knowledge retrieval and candidate pre-selection to guide the LLM API, KGST achieves more accurate and reliable STL generation compared to purely grammar-guided approaches. RESTL further incorporates reinforcement learning into NL-to-STL generation through multi-aspect rewards, curriculum learning, and PPO-based optimization~\cite{fang2026restl}. These methods improve final-formula prediction and candidate selection, but their supervision remains primarily centered on the final STL output rather than on explicit intermediate computations.

Recent work has also explored LLM-based NL-to-STL generation, including LLaMA-style models for translating natural-language requirements into STL formulas~\cite{mao2024nl2stl}. These studies suggest that large language models can serve as useful backbones for STL specification generation. However, existing NL-to-STL methods still underrepresent realistic CPS requirements involving domain-grounded signals, bilingual descriptions, physical units, arithmetic reasoning, nested temporal structures, event-triggered semantics, and verifiable intermediate computation. \textsc{ReasonSTL} addresses these aspects through a local tool-augmented drafting workflow, outcome-bounded process supervision, and a bilingual computation-aware benchmark with structured STL trees and tool-use annotations.

\section{Additional STL Definitions}
\label{app:stl_details}

This section provides additional STL constructs used in \textsc{STL-Bench}. Beyond standard future-time STL operators, the benchmark includes event predicates and past-time operators to represent event-triggered and history-dependent requirements that commonly arise in cyber--physical systems.

\paragraph{Rise and fall predicates.}
Given an atomic predicate $\pi^\mu$, where $\pi^\mu[k]$ holds iff $\mu(s_k)\ge 0$, the derived event predicates \emph{rise} and \emph{fall} are defined as
\begin{align}
    \mathrm{rise}(\pi^\mu,k)
    &:= \neg \pi^\mu[k-1] \wedge \pi^\mu[k]
    \equiv (\mu(s_{k-1}) < 0) \wedge (\mu(s_k) \ge 0), \\
    \mathrm{fall}(\pi^\mu,k)
    &:= \pi^\mu[k-1] \wedge \neg \pi^\mu[k]
    \equiv (\mu(s_{k-1}) \ge 0) \wedge (\mu(s_k) < 0).
\end{align}
Thus, $\mathrm{rise}(\pi^\mu,k)$ captures a threshold-crossing event from false to true at time $k$, whereas $\mathrm{fall}(\pi^\mu,k)$ captures the corresponding transition from true to false. These predicates are useful for requirements involving triggering conditions, such as a signal rising above an operational threshold or falling below a safety limit.

\paragraph{Historically and once operators.}
We also use the past-time operator $\mathbf{H}_{[a,b]}$, known as \emph{historically}, to express requirements over a finite history window:
\begin{equation}
    (\mathbf{s},k)\models \mathbf{H}_{[a,b]}\phi
    \quad \mathrm{iff} \quad
    \forall k'\in[k-b,k-a],\;(\mathbf{s},k')\models\phi .
\end{equation}
That is, $\mathbf{H}_{[a,b]}\phi$ holds at time $k$ if $\phi$ has continuously held over the past interval $[k-b,k-a]$. We further include the derived past-time operator $\mathbf{O}_{[a,b]}$, known as \emph{once}, defined as
\begin{equation}
    (\mathbf{s},k)\models \mathbf{O}_{[a,b]}\phi
    \quad \mathrm{iff} \quad
    \exists k'\in[k-b,k-a],\;(\mathbf{s},k')\models\phi .
\end{equation}
Together, \emph{rise}, \emph{fall}, $\mathbf{H}$, and $\mathbf{O}$ allow \textsc{STL-Bench} to include event-based and history-dependent specifications beyond standard future-time STL patterns.

\newpage
\section{Structured JSON Schema}
\label{app:json_schema}

\textsc{ReasonSTL} represents STL formulas as structured JSON trees rather than flat strings. The root node specifies the top-level STL operator, internal nodes encode temporal or Boolean composition, and atomic predicates appear as normalized leaf strings. This representation makes the formula hierarchy explicit and avoids ambiguities caused by parentheses, operator precedence, whitespace, and surface-level formatting. It also provides a unified interface for schema validation, recursive tree matching, subformula-level reward computation, and diagnostic analysis.

The following example illustrates the JSON-tree representation used in \textsc{STL-Bench}:
\begin{verbatim}
{
  "STL": {
    "Operation": "Finally",
    "Time": [0, 1800],
    "Leftaction": null,
    "Rightaction": {
      "Operation": "and",
      "SubQueries": [
        "altitude>3048.0",
        "speed>=102.89"
      ]
    }
  }
}
\end{verbatim}

This JSON tree corresponds to the STL formula
\begin{equation}
    \mathbf{F}_{[0,1800]}
    \big(
    (\mathrm{altitude}>3048.0)
    \wedge
    (\mathrm{speed}\ge 102.89)
    \big).
\end{equation}
The field \texttt{Operation} specifies the operator type, \texttt{Time} specifies the temporal interval when applicable, and \texttt{SubQueries} stores the children of Boolean operators. Unary temporal operators such as \texttt{Finally}, \texttt{Globally}, \texttt{Historically}, and \texttt{Once} use \texttt{Rightaction} to store their child formula. Binary temporal operators such as \texttt{Until} and \texttt{Since} use \texttt{Leftaction} and \texttt{Rightaction} to store the left and right subformulas. Boolean operators such as \texttt{and} and \texttt{or} use \texttt{SubQueries} to support multi-branch formulas. Atomic predicates are represented as normalized strings with canonical signal names, comparison operators, and numerical thresholds.

\begin{table}[h]
\centering
\footnotesize
\caption{Main fields in the structured STL JSON schema.}
\label{tab:json_schema_fields}
\setlength{\tabcolsep}{4pt}
\renewcommand{\arraystretch}{1.08}
\begin{tabular}{@{}p{0.22\linewidth}p{0.25\linewidth}p{0.45\linewidth}@{}}
\toprule
\textbf{Field} & \textbf{Type} & \textbf{Description} \\
\midrule
\texttt{STL} & object & Root field containing the full STL tree \\
\texttt{Operation} & string & Operator name, e.g., \texttt{Finally}, \texttt{Globally}, \texttt{Until}, \texttt{and} \\
\texttt{Time} & list or null & Temporal interval $[a,b]$ for bounded temporal operators \\
\texttt{Leftaction} & object/string/null & Left child for binary temporal or implication-like operators \\
\texttt{Rightaction} & object/string/null & Child formula for unary operators, or right child for binary operators \\
\texttt{SubQueries} & list & Children of multi-branch Boolean operators such as \texttt{and} and \texttt{or} \\
Predicate leaf & string & Normalized atomic predicate, e.g., \texttt{altitude>3048.0} \\
\bottomrule
\end{tabular}
\end{table}

Before evaluation, each prediction is parsed and canonicalized. Canonicalization includes normalizing operator names, checking required fields, converting numerical fields into a standard format, validating temporal intervals, and parsing predicate strings into signal--operator--threshold triples. If the prediction cannot be parsed into a valid JSON tree, it is assigned zero format reward and zero tree-matching score. Otherwise, the parsed tree is used for both automatic evaluation and reward computation.

The recursive tree matcher compares the predicted tree $\hat{\phi}$ with the reference tree $\phi^\star$ from the root downward. Operator nodes must match in type, temporal intervals are compared with a fixed numerical tolerance, and predicate leaves are compared by signal name, comparison direction, and threshold value. For commutative Boolean operators such as \texttt{and} and \texttt{or}, child order is treated as invariant: the matcher searches for the best alignment between predicted and reference children before computing the subtree score. This avoids penalizing formulas that differ only by the ordering of conjuncts or disjuncts.

Formally, the matcher returns a similarity score
\begin{equation}
    S^{\mathrm{tree}}(\hat{\phi},\phi^\star)\in[0,1],
\end{equation}
where $1$ indicates an exact canonical tree match. The matcher assigns credit in a top-down manner: a node contributes to the score only if its operator type or predicate form matches the corresponding reference node, and child nodes are evaluated only under matched parent nodes. Consequently, if an internal operator is incorrect, the entire subtree rooted at that node receives no further credit, even if some descendant predicates happen to match the reference. For example, a correct threshold appearing under an incorrect temporal operator is not rewarded as an independently correct leaf. This design encourages models to recover the intended hierarchical STL structure rather than only matching isolated predicate strings. The resulting score is used in the content reward described in Section~\ref{sec:method_reward}. To preserve a strict preference for exact formal correctness, non-perfect tree matches are capped by the factor $\kappa<1$, so that partially correct trees can provide informative learning signals but cannot receive the full content reward.

This structured representation is also used to compute the automatic metrics in Section~\ref{sec:nl2stl_task}. Format Accuracy checks whether the model output can be parsed into the required schema and, when evaluating formula skeletons, whether the masked tree structure matches the reference. Formula Accuracy further requires the complete canonical tree, including predicates, temporal intervals, and numerical thresholds, to match the reference under the tree-matching protocol. Thus, the JSON schema serves not only as an output format, but also as the central representation for validation, evaluation, reward design, and error analysis.

\section{Additional Details of \textsc{STL-Bench}}
\label{app:bench_details}

This section provides additional details on the construction of \textsc{STL-Bench}, complementing the benchmark overview in Section~\ref{sec:bench}. We describe the domain taxonomy, template design, model-assisted requirement realization, tool-use annotations, validation and deduplication procedures, split construction, and human auditing protocol.

\subsection{Domain and Scenario Taxonomy}

\textsc{STL-Bench} is organized around six engineering domains: autonomous driving, robotics, industrial control, environmental monitoring, electrical systems, and aerospace systems. Each domain is associated with concrete CPS scenarios and semantically meaningful signal variables. This design encourages models to ground natural-language requirements in physically interpretable predicates rather than abstract symbols. In total, the benchmark covers 33 scenarios and 41 canonical signal variables.

\begin{table}[h]
\centering
\footnotesize
\setlength{\tabcolsep}{4pt}
\renewcommand{\arraystretch}{1.08}
\caption{Domain and scenario taxonomy of \textsc{STL-Bench}.}
\label{tab:domain_taxonomy}
\begin{tabular}{@{}p{0.18\linewidth}p{0.47\linewidth}p{0.28\linewidth}@{}}
\toprule
\textbf{Domain} & \textbf{Scenarios} & \textbf{Example Signal Variables} \\
\midrule
Autonomous driving
& AEB, ACC, lane keeping, parking, fuel monitoring, traction control
& velocity, acceleration, steering, brake, obstacle distance \\
Robotics
& pick-and-place, welding, collision avoidance, assembly, mobile navigation
& position, distance, torque, contact force, joint velocity \\
Industrial control
& reactor control, CNC machining, conveyor belt, hydraulic press, boiler system, structural monitoring
& pressure, temperature, flow rate, load, strain, stress \\
Environmental monitoring
& indoor climate, water quality, air quality, greenhouse, noise monitoring
& humidity, CO$_2$ level, pollutant concentration, pH level, noise level \\
Electrical systems
& battery management, motor drive, power grid, solar panel, signal processing
& voltage, current, frequency, power, phase, amplitude \\
Aerospace systems
& altitude hold, takeoff landing, drone survey, satellite attitude, flight envelope, UAV delivery
& altitude, airspeed, pitch, roll, yaw, heading \\
\bottomrule
\end{tabular}
\end{table}

The canonical signal vocabulary is shared across English and Chinese requirements:
\begin{equation}
\begin{aligned}
\mathcal{V}=\{&
\texttt{accel},\texttt{acceleration},\texttt{altitude},\texttt{amplitude},\texttt{brake},
\texttt{brightness},\texttt{co2\_level},\texttt{concentration},\\
&\texttt{current},\texttt{density},\texttt{dist},\texttt{distance},\texttt{flow\_rate},
\texttt{frequency},\texttt{fuel\_level},\texttt{heading},\\
&\texttt{humidity},\texttt{load},\texttt{noise\_level},\texttt{oxygen},\texttt{ph\_level},
\texttt{phase},\texttt{pitch},\texttt{power},\\
&\texttt{pressure},\texttt{roll},\texttt{rpm},\texttt{speed},\texttt{steering},
\texttt{strain},\texttt{stress},\texttt{temp},\\
&\texttt{temperature},\texttt{throttle},\texttt{torque},\texttt{velocity},
\texttt{voltage},\texttt{x\_pos},\texttt{y\_pos},\texttt{yaw},\texttt{z\_pos}, \dots \}.
\end{aligned}
\end{equation}

\subsection{Template and Complexity Design}

Samples are generated from scenario-conditioned STL templates. Each template specifies admissible signal variables, numerical ranges, temporal intervals, operator structures, optional unit systems, and optional tool-use requirements. The supported operators include \texttt{Globally}, \texttt{Finally}, \texttt{Until}, \texttt{Since}, \texttt{imply}, \texttt{and}, \texttt{or}, \texttt{Not}, \texttt{Rise}, \texttt{Fall}, \texttt{Historically}, and \texttt{Once}. Compared with a single grammar-only generator, the scenario-conditioned design provides tighter control over domain semantics, physically plausible value ranges, and computation-sensitive fields.

\begin{table}[h]
\centering
\footnotesize
\setlength{\tabcolsep}{4pt}
\renewcommand{\arraystretch}{1.08}
\caption{Complexity levels used in \textsc{STL-Bench}.}
\label{tab:complexity_levels}
\begin{tabular}{@{}cp{0.65\linewidth}c@{}}
\toprule
\textbf{Level} & \textbf{Description} & \textbf{Sampling Ratio} \\
\midrule
1 & Single operator with one atomic predicate & 25\% \\
2 & Two operators with two to three predicates & 25\% \\
3 & Three operators with three to five predicates & 20\% \\
4 & Multi-stage temporal constraints with four to six predicates & 15\% \\
5 & Deeply nested multi-branch formulas with five to seven predicates & 10\% \\
6 & Highly complex formulas with six to eight predicates and chained tool reasoning & 5\% \\
\bottomrule
\end{tabular}
\end{table}

The complexity level is determined by formula depth, number of temporal or Boolean operators, number of atomic predicates, and length of the tool-use chain. Levels 5--6 are designed to stress nested temporal structures, multi-signal dependencies, and chained intermediate computations.

\subsection{Model-Assisted Requirement Realization}

The formal label is determined before natural-language realization. For each candidate sample, the template first fixes the scenario, signal variables, operator structure, temporal intervals, thresholds, and tool-use requirements. A strong language model is then used only to realize this fixed formal specification as an engineering-style natural-language requirement. This separation reduces semantic drift because the reference STL tree is not inferred from generated text after the fact.

We use DeepSeek-Chat with decoding temperature $0.9$. Each API call generates a small batch of candidate requirements from a structured prompt containing the domain description, scenario, admissible signal variables, signal ranges, target complexity level, STL JSON schema, tool-call format, and quality constraints. A candidate is retained only if it passes automatic validation, including natural-language length checks, STL-tree validity, signal-name validity, tool-call executability, and consistency between tool outputs and final STL fields.

For bilingual construction, English and Chinese requirements are generated independently under the same scenario and complexity conditions, rather than by direct translation. This preserves a shared formal schema and signal vocabulary while allowing language-specific phrasing, domain terminology, and engineering style.

\subsection{Tool-Use Annotation}

For computation-sensitive samples, \textsc{STL-Bench} provides explicit tool-use annotations. Each annotation records the tool name, input arguments, expected output, and the target field in the final STL JSON where the computed value is used. The tool set includes:
\begin{itemize}
    \item \texttt{parse\_duration}: normalizes natural-language duration expressions into canonical time units;
    \item \texttt{convert\_unit}: converts physical quantities into canonical units;
    \item \texttt{eval\_math\_expr}: evaluates arithmetic expressions used in thresholds or time bounds;
    \item \texttt{calc\_time\_diff}: computes time differences from temporal expressions or timestamps.
\end{itemize}

\begin{table}[h]
\centering
\footnotesize
\setlength{\tabcolsep}{4pt}
\renewcommand{\arraystretch}{1.08}
\caption{Examples of tool-use annotations.}
\label{tab:tool_examples}
\begin{tabular}{@{}p{0.19\linewidth}p{0.43\linewidth}p{0.13\linewidth}p{0.18\linewidth}@{}}
\toprule
\textbf{Tool} & \textbf{Input} & \textbf{Output} & \textbf{Used in STL Field} \\
\midrule
\texttt{parse\_duration}
& ``30 minutes''
& 1800
& Time \\
\texttt{convert\_unit}
& $(10000,\mathrm{ft},\mathrm{m})$
& 3048.0
& predicate threshold \\
\texttt{convert\_unit}
& $(200,\mathrm{kn},\mathrm{m/s})$
& 102.89
& predicate threshold \\
\texttt{eval\_math\_expr}
& \{"expression": "2*900"\}
& 1800
& time bound \\
\texttt{calc\_time\_diff}
& ``time interval between 2025-08-01 8:00:00 and 2025-08-01 8:15:00''
& 900
& Time \\
\bottomrule
\end{tabular}
\end{table}

\subsection{Event Semantics}

We explicitly distinguish threshold-state requirements from edge-triggered event requirements. Surface expressions such as ``rises above'', ``drops below'', or ``surges past'' are not automatically mapped to \texttt{Rise} or \texttt{Fall}. They are annotated as edge events only when the requirement specifies a transition from violation to satisfaction, or from satisfaction to violation. Otherwise, they are treated as ordinary threshold predicates. This distinction is important because an edge-triggered event and a threshold-state predicate can have different STL semantics even when their natural-language descriptions share similar lexical cues.

\subsection{Validation and Deduplication}

All generated samples are processed by automatic validators before inclusion. The validators check natural-language length constraints, JSON syntax, required fields, STL-tree well-formedness, operator arity, temporal-interval validity, predicate format, signal-variable validity, tool-call executability, unit-conversion correctness, arithmetic correctness, and consistency between tool outputs and final STL values.

Invalid samples are removed or repaired only when the correction is deterministic. For example, malformed atomic predicates containing conjunctions or disjunctions can be converted into Boolean subtrees, while unexecutable tool calls or inconsistent tool outputs lead to sample removal. This avoids introducing subjective corrections into the formal labels.

Deduplication is performed in two stages. First, exact and symbolic duplicates are removed using normalized natural-language strings, STL-structure hashing, and predicate assignments. Second, embedding-based semantic pruning is applied to reduce near-duplicate natural-language realizations. Requirements are encoded with Qwen3-Embedding-4B, samples with cosine similarity above $0.9$ are clustered, and each cluster retains samples with higher formula complexity, clearer domain grounding, or better linguistic quality.

\subsection{Data Splits and Manual Test Set}

The standard train, validation, and test sets are constructed after deduplication using stratified random partitioning over language, domain, complexity level, and tool-use type. This preserves the overall benchmark distribution while reducing leakage from near-duplicate requirements. The split procedure uses a fixed random seed.

In addition to the standard split, \textsc{STL-Bench} includes a scenario-aware held-out split, where selected scenarios from each high-level domain are excluded from training and used only for validation or testing. This split evaluates whether models can generalize across scenario variations rather than merely memorizing domain--template combinations.

We also construct an additional manually written test set of approximately 200 requirements. These samples are authored by human annotators in English and Chinese without using the data-generation templates, paired with manually specified structured STL formulas, and checked by the same validation pipeline. This set is used to assess generalization beyond the template-anchored generation process.

\subsection{Human Auditing Protocol}

A stratified subset of samples is manually audited to assess dataset quality. The audit checks natural-language/STL semantic alignment, JSON validity, tool-result correctness, bilingual terminology consistency, formula readability, and the distinction between threshold-state and edge-triggered requirements. The audited subset is sampled across languages, domains, complexity levels, and tool-use types.

\textsc{STL-Bench} is not exhaustively verified sample by sample. Instead, it combines template-controlled label construction, deterministic validation, symbolic filtering, embedding-based semantic pruning, and stratified human auditing. This design balances dataset scale with formal-label reliability while leaving room for residual ambiguity or annotation noise, as discussed in Section~\ref{app:limitations}.

\section{Additional Method Details}
\label{app:method_details}

\subsection{Tool Execution Protocol}

\textsc{ReasonSTL} uses an explicit tool-execution interface to separate semantic interpretation from deterministic computation. During generation, the model may emit a \texttt{<tool\_call>} block with a tool name and structured arguments. If the call is syntactically valid and belongs to the predefined tool set, the environment executes the tool and appends the returned value as a \texttt{<tool\_result>} block. The model then continues generation conditioned on the original requirement, previous reasoning, and the verified tool output.

Consistent with Section~\ref{sec:method_framework}, a rollout is represented as
\begin{equation}
    y=(z_1,u_1,o_1,\ldots,z_M),
\end{equation}
where $z_m$ is a model-generated segment, $u_m$ is an optional tool call, and $o_m$ is the deterministic tool output. Since tool-output tokens are produced by the execution environment rather than sampled from the model policy, they are excluded from policy-gradient updates by the token mask $M_{i,t}$.

\begin{verbatim}
<think>
The requirement uses "30 minutes", which should be converted to seconds.
</think>
<tool_call>
parse_duration("30 minutes")
</tool_call>
<tool_result>
1800
</tool_result>
\end{verbatim}

Generation terminates when the model emits the final STL JSON answer, reaches the maximum number of tool rounds, exceeds the maximum generation length, or violates the reasoning-length constraint. Malformed tool calls, unsupported tools, invalid arguments, failed execution, or tool outputs inconsistent with the final formula mark the corresponding intermediate stage as invalid.

\subsection{Stage Validation Details}

Each rollout is parsed into intermediate stages and a final STL-construction stage. Intermediate stages include reasoning segments, tool calls, tool outputs, and the subsequent use of computed values. The deterministic validator assigns a binary correctness score $c_{i,k}\in\{0,1\}$ to each intermediate stage. Table~\ref{tab:stage_validation} summarizes the main validation checks.

\begin{table}[h]
\centering
\footnotesize
\caption{Validation checks for intermediate stages.}
\label{tab:stage_validation}
\setlength{\tabcolsep}{4pt}
\renewcommand{\arraystretch}{1.08}
\begin{tabular}{@{}P{0.20\linewidth}P{0.38\linewidth}P{0.34\linewidth}@{}}
\toprule
\textbf{Stage Type} & \textbf{Check} & \textbf{Example} \\
\midrule
Tool selection & Tool name is valid and appropriate & \texttt{parse\_duration} for ``30 minutes'' \\
Argument format & Arguments satisfy the tool schema & Valid unit string or arithmetic expression \\
Execution & Tool can be executed successfully & No parsing or conversion failure \\
Result consistency & Tool output matches later formula fields & $30$ min $\rightarrow 1800$ sec in interval bound \\
Predicate grounding & Signal and threshold match requirement & Correct variable and inequality direction \\
Event grounding & Edge operator is semantically justified & \emph{rise}/\emph{fall} only for explicit transitions \\
\bottomrule
\end{tabular}
\end{table}

For event-related predicates, surface expressions such as ``rises above'', ``drops below'', or ``surges past'' are not sufficient to trigger \emph{rise} or \emph{fall}. These operators are considered valid only when the requirement explicitly describes a transition from violation to satisfaction, or vice versa.

\subsection{Tree Matching and Numerical Tolerance}

The final STL JSON is evaluated by the recursive tree matcher described in Section~\ref{sec:method_reward}. If the output cannot be parsed into a valid JSON tree, we set $R_i^{\mathrm{fmt}}=0$ and $S_i^{\mathrm{tree}}=0$. Otherwise, the matcher recursively compares operators, temporal intervals, predicates, and subformulas. Commutative Boolean operators are matched in an order-invariant manner, and numerical fields use a fixed tolerance of $0.1$ for temporal bounds and predicate thresholds. Non-perfect matches are capped by the content cap $\kappa$, as defined in Eq.~\eqref{eq:capped_tree_reward}.

\subsection{Stage Alignment and Token Masking}

For group-relative advantage estimation, final outcomes are normalized against final outcomes, while intermediate stages are normalized against corresponding intermediate stages. In practice, intermediate stages are aligned by their parsed role, such as duration parsing, unit conversion, arithmetic evaluation, event grounding, or final-value incorporation. This avoids mixing structurally different rewards when computing $A_{i,k}^{\mathrm{proc}}$.

The token mask $M_{i,t}$ excludes four types of tokens: invalid tokens, truncated tokens, deterministic tool-output tokens, and tokens belonging to process stages blocked by the prefix mask in Eq.~\eqref{eq:effective_process_reward}. Thus, only model-generated tokens with well-defined stage-level credit contribute to the policy update.

\subsection{Optimization Procedure}

Algorithm~\ref{alg:app_process_reward} summarizes the outcome-bounded process-rewarded optimization procedure. It follows the notation and reward definitions in Section~\ref{sec:method_reward}.

\begin{algorithm}[h]
\caption{Outcome-Bounded Process-Rewarded Optimization}
\label{alg:app_process_reward}
\begin{algorithmic}[1]
\Require Requirement $q$, old policy $\pi_{\theta_{\mathrm{old}}}$, group size $G$
\State Sample rollouts $\{y_i\}_{i=1}^{G}\sim \pi_{\theta_{\mathrm{old}}}(\cdot\mid q)$
\For{$i=1,\ldots,G$}
    \State Parse $y_i$ into intermediate stages and final STL tree $\hat{\phi}_i$
    \State Validate intermediate stages to obtain $\{c_{i,k}\}_{k=1}^{K_i}$
    \State Compute final outcome reward $R_i^{\mathrm{out}}$ and correctness indicator $C_i^{\mathrm{final}}$
    \For{$k=1,\ldots,K_i$}
        \State Compute process reward $R_{i,k}^{\mathrm{proc}}$ using Eq.~\eqref{eq:effective_process_reward}
    \EndFor
\EndFor
\State Compute outcome advantages $\{A_i^{\mathrm{out}}\}_{i=1}^{G}$
\State Compute process advantages $\{A_{i,k}^{\mathrm{proc}}\}$ over aligned stages
\State Assign token-level advantages $A_{i,t}$ and masks $M_{i,t}$
\State Update $\pi_\theta$ using the objective in Eq.~\eqref{eq:process_objective}
\end{algorithmic}
\end{algorithm}

\subsection{SFT Cold Start and RL Optimization}

For \textsc{STL-Bench}, \textsc{ReasonSTL} uses a short SFT cold start before reinforcement learning. The SFT stage initializes the model with the structured JSON output format and the basic think--tool interaction pattern. This improves early rollout quality by increasing the probability of parseable JSON, valid tool-call syntax, and recoverable intermediate traces. The final performance is then improved through tool-augmented reinforcement learning with outcome-bounded process rewards. Section~\ref{app:sft_rl_ablation} provides controlled comparisons among SFT-only, RL-only, and SFT-initialized RL variants.

\subsection{Training Hyperparameters}
\begin{table}[h]
\centering
\footnotesize
\caption{Training hyperparameters for \textsc{ReasonSTL}.}
\label{tab:hyperparams_appendix}
\setlength{\tabcolsep}{4pt}
\renewcommand{\arraystretch}{1.08}
\begin{tabular}{@{}P{0.42\linewidth}P{0.50\linewidth}@{}}
\toprule
\textbf{Hyperparameter} & \textbf{Value} \\
\midrule
Backbone & Qwen3-4B \\
Fine-tuning method & Full fine-tuning, with input embeddings frozen \\
Training GPUs & $8\times$H20 \\
Evaluation GPU & $1\times$H20 \\
\midrule
SFT learning rate & $2\times 10^{-5}$ \\
SFT per-GPU batch size & 2 \\
SFT gradient accumulation & 4 \\
SFT effective batch size & 64 \\
SFT optimizer & AdamW, weight decay 0.01 \\
SFT warmup ratio & 0.05 \\
SFT max sequence length & 2048 \\
SFT epochs & 1 \\
SFT curriculum & First 30\% steps use samples with tree depth $\leq 3$ \\
\midrule
RL learning rate & $3\times 10^{-6}$ \\
RL LR schedule & Warmup + cosine decay, minimum ratio 0.1 \\
RL warmup steps & 20 \\
RL batch size & 16 \\
RL micro batch size & 2 \\
Group size $G$ & 8 \\
Maximum tool rounds & 5 \\
Maximum generation tokens & 2048 \\
Training / test temperature & 1.0 / 0.0 \\
Top-$p$ & 0.95 \\
RL optimizer & AdamW \\
Gradient clipping & 1.0 \\
RL epochs & 9 \\
\midrule
Outcome-bound factor $\tau$ & 0.5 \\
Content partial cap $\kappa$ & 0.3 \\
Numerical tolerance & 0.1 \\
\midrule
Direct RL steps without think/tool & 3K \\
Full RL steps & 5K \\
Direct training time & $\sim$20 hours \\
Full training time & $\sim$5 days \\
\bottomrule
\end{tabular}
\end{table}

The implementation uses an unclipped group-relative objective with token-level stage advantages. Stability is controlled by group-normalized advantages, outcome-bounded process rewards, prefix masking, gradient clipping, and the SFT cold start. Since the objective does not use PPO-style likelihood-ratio clipping, no clipping coefficient is introduced.

\section{Additional Experiments}
\label{app:additional_experiments}

\subsection{SFT and RL Ablation}
\label{app:sft_rl_ablation}

We conduct ablations to disentangle the effects of supervised cold-start training, reinforcement learning, think--tool interaction, and process-level rewards. The compared variants include pure SFT, RL from the base model, and SFT-initialized RL. For RL-based variants, we distinguish direct RL, which optimizes final-answer correctness without explicit think--tool interaction, from process-rewarded RL, which additionally supervises intermediate tool-use stages.

\begin{table}[h]
\centering
\footnotesize
\setlength{\tabcolsep}{13pt}
\renewcommand{\arraystretch}{1.08}
\caption{
SFT/RL ablation on \textsc{STL-Bench}. Results are reported as Formula Accuracy / Format Accuracy, together with SFT and RL training steps.
}
\label{tab:sft_rl_ablation}
\begin{tabular}{@{}lcccccc@{}}
\toprule
\textbf{Variant}
& \textbf{Think}
& \textbf{Tool}
& \textbf{SFT}
& \textbf{RL}
& \textbf{English}
& \textbf{Chinese} \\
\midrule
Qwen3-4B
& \xmark & \xmark
& 0
& 0
& 0.070 / 0.110
& 0.090 / 0.130 \\
SFT-only Direct
& \xmark & \xmark
& 2K
& 0
& 0.200 / 0.240
& 0.240 / 0.280 \\
RL-only Direct
& \xmark & \xmark
& 0
& 5.5K
& 0.410 / 0.510
& 0.330 / 0.380 \\
RL-only Think+Tool
& \cmark & \cmark
& 0
& 8K
& 0.500 / 0.560
& 0.460 / 0.470 \\
SFT + Direct RL
& \xmark & \xmark
& 300
& 3K
& 0.420 / 0.500
& 0.330 / 0.410 \\
\textbf{ReasonSTL}
& \cmark & \cmark
& 300
& 5K
& \textbf{0.510 / 0.560}
& \textbf{0.470 / 0.490} \\
\bottomrule
\end{tabular}
\end{table}

Table~\ref{tab:sft_rl_ablation} shows that SFT alone improves over the base Qwen3-4B model but remains substantially below RL-based variants. This suggests that supervised imitation helps the model acquire the target output format, but is insufficient for robust computation-aware STL construction. Direct RL substantially improves both Formula Accuracy and Format Accuracy, indicating that reward-based optimization is important for adapting the model to the structured evaluation objective.

The comparison between RL-only and SFT-initialized RL indicates that the SFT cold start mainly improves training efficiency and stability. With only 300 SFT steps, SFT + Direct RL reaches performance comparable to RL-only Direct while using fewer RL steps. In the tool-augmented setting, RL-only Think+Tool already achieves strong performance, while SFT + Process-Rewarded RL obtains the best overall results with fewer RL steps than RL-only Think+Tool. These results support the use of SFT as a lightweight initialization strategy and indicate that the final performance gains primarily arise from reinforcement learning, explicit think--tool interaction, and process-level supervision.

\subsection{Representation Study: Structured JSON vs. Flat String}

We compare structured JSON representation with flat string-form STL generation under the same GPT-based API baseline. As shown in Table~\ref{tab:json_vs_str}, JSON mode does not necessarily improve exact Formula Accuracy for a prompted black-box model, but it yields higher Template Accuracy, suggesting that the structured representation better preserves the high-level operator skeleton. More importantly, JSON provides a stable interface for trainable local models: the output schema is easier to validate, syntactic errors are easier to detect, and partial rewards can be assigned through recursive tree matching. Flat STL strings are more sensitive to parentheses, operator precedence, and surface formatting, making automatic evaluation and reward design less reliable. We therefore use structured JSON as the default representation for \textsc{STL-Bench} and \textsc{ReasonSTL}.

\begin{table}[h]
\centering
\small
\caption{Comparison between structured JSON and flat string STL representations under the same GPT-based API baseline.}
\label{tab:json_vs_str}
\begin{tabular}{lcc}
\toprule
\textbf{Metric} & \textbf{JSON Mode} & \textbf{String Mode} \\
\midrule
Formula Accuracy & 20.5\% & 22.8\% \\
Template Accuracy & 31.3\% & 24.0\% \\
\bottomrule
\end{tabular}
\end{table}

\subsection{Error Breakdown}

We categorize incorrect predictions to identify the dominant failure modes of prompted API models on \textsc{STL-Bench}. The analysis focuses on whether failures arise from format control, temporal-operator grounding, predicate grounding, numerical computation, or tool-use behavior.

\begin{table}[h]
\centering
\small
\caption{
Major error types of GPT-5.4 on \textsc{STL-Bench}. Percentages are computed over 75 incorrect predictions after excluding reference-quality issues. Categories are not mutually exclusive.
}
\label{tab:error_breakdown_appendix}
\begin{tabular}{lcc}
\toprule
\textbf{Error Type} & \textbf{Count} & \textbf{Error Share} \\
\midrule
Temporal-operator hallucination & 30 & 40.0\% \\
Numerical / unit-conversion error & 26 & 34.7\% \\
Signal-name ambiguity & 26 & 34.7\% \\
\bottomrule
\end{tabular}
\end{table}

The most frequent error is temporal-operator hallucination: models introduce formal event predicates such as \emph{rise} or \emph{fall} from surface cues such as ``rises above'' or ``drops below'', even when the requirement describes ordinary threshold satisfaction. Numerical and unit-conversion errors motivate deterministic tool use, while signal-name ambiguity motivates signal-alias normalization and human verification.

\subsection{Qualitative Error Cases}

We present qualitative examples to complement the aggregate metrics and error statistics. These cases illustrate systematic failures of black-box API models, remaining limitations of \textsc{ReasonSTL}, and cases where strict automatic matching either underestimates semantically plausible predictions or reveals residual reference-quality issues.

\paragraph{API failure: temporal-operator hallucination.}
A common failure mode of API-based models is to over-interpret surface-level motion verbs as formal STL event predicates. Consider the requirement:
\begin{quote}
\emph{If the current surges past 75 amperes, the system must historically have had stable voltage.}
\end{quote}
The reference formula treats the antecedent as a threshold-state predicate:
\begin{equation}
    \mathrm{imply}(\mathrm{current} > 75,\ldots).
\end{equation}
In contrast, GPT-5.4 predicts:
\begin{equation}
    \mathrm{imply}(\mathrm{Rise}(\mathrm{current} > 75),\ldots).
\end{equation}
This prediction is semantically incorrect: ``surges past'' describes threshold exceedance, but does not necessarily specify a formal edge-triggered transition. In STL, \emph{Rise} denotes a transition from predicate violation to predicate satisfaction, which is stronger than ordinary threshold satisfaction and changes the requirement semantics.

\paragraph{Reference-quality issue: incorrect geometric annotation.}
Manual inspection also reveals cases where automatic mismatches are caused by reference annotation errors rather than model errors. Consider:
\begin{quote}
\emph{Throughout the placement segment from time 20 to 80 seconds, the end-effector must stay within a 5 cm radius of the target $(X=1800\mathrm{mm},Y=200\mathrm{mm})$ whenever its $Z$ position is less than 10 cm.}
\end{quote}
The reference formula encodes the geometric condition as axis-aligned bounds:
\begin{equation}
\begin{aligned}
\mathbf{G}_{[20,80]}\big(
\mathrm{imply}(&\mathrm{z\_pos}<0.1,\;
(\mathrm{x\_pos}>1.795)\wedge(\mathrm{x\_pos}<1.805)\\
&\wedge(\mathrm{y\_pos}>0.195)\wedge(\mathrm{y\_pos}<0.205))
\big).
\end{aligned}
\end{equation}
However, the natural language specifies a radius constraint around the target point, which is more directly represented as:
\begin{equation}
    \mathbf{G}_{[20,80]}
    \big(
    \mathrm{imply}
    (
    \mathrm{z\_pos}<0.1,\;
    \sqrt{(\mathrm{x\_pos}-1.8)^2+(\mathrm{y\_pos}-0.2)^2}\le 0.05
    )
    \big).
\end{equation}
The reference bounds correspond to a substantially narrower axis-aligned tolerance, suggesting a conversion or template-instantiation error. This case illustrates that, although \textsc{STL-Bench} uses template control, rule-based validation, semantic deduplication, and stratified auditing, residual label errors can remain in scalable benchmark construction.

\paragraph{\textsc{ReasonSTL} failure: temporal scope error.}
\textsc{ReasonSTL} can still fail on nested temporal scope. Consider:
\begin{quote}
\emph{During the 120-second antenna pointing maneuver, the pitch must stay under 1.2 degrees, and if it exceeds 0.8 degrees, it must fall back below 0.5 degrees within 15 seconds.}
\end{quote}
The reference formula is:
\begin{equation}
    \mathbf{G}_{[0,120]}
    \big(
        (\mathrm{pitch}<1.2)
        \wedge
        \mathrm{imply}
        (
            \mathrm{pitch}>0.8,\;
            \mathbf{F}_{[0,15]}(\mathrm{pitch}<0.5)
        )
    \big).
\end{equation}
\textsc{ReasonSTL} predicts:
\begin{equation}
    \mathbf{G}_{[0,120]}(\mathrm{pitch}<1.2)
    \wedge
    \mathrm{imply}
    (
        \mathrm{pitch}>0.8,\;
        \mathbf{F}_{[0,15]}(\mathrm{pitch}<0.5)
    ).
\end{equation}
The prediction incorrectly lifts the implication outside the global temporal scope. Consequently, the recovery constraint is no longer enforced throughout the 120-second maneuver; it is evaluated only at the top level. This example shows that preserving temporal scope remains challenging for nested STL formulas.

\paragraph{Automatic failure but human-verified success: signal-name ambiguity.}
Strict automatic matching can reject semantically plausible predictions when the model uses a different canonical signal name. For a requirement referring to ``rotational speed'', the reference predicate may use
\begin{equation}
    \mathrm{rpm} > X,
\end{equation}
whereas model may predict
\begin{equation}
    \mathrm{rotational\_speed} > X.
\end{equation}
The prediction is marked incorrect by exact tree matching because the variable names differ, but the predicted signal name is semantically aligned with the requirement. This motivates human verification, signal-alias normalization, and semantic-equivalence-aware evaluation.

\section{Limitations}
\label{app:limitations}

Several limitations remain. First, \textsc{STL-Bench} is not exhaustively verified by human annotators. Although its construction combines template-controlled formula generation, rule-based validation, symbolic filtering, embedding-based deduplication, and stratified human auditing, not every individual sample is manually checked. As a result, residual annotation errors, ambiguous requirement interpretations, or imperfect natural-language/STL alignments may still exist.

Second, a natural-language requirement does not always correspond to a unique STL formula. Multiple specifications may be semantically reasonable depending on modeling conventions, signal abstractions, temporal granularity, and whether auxiliary assumptions are made explicit. This work favors compact and canonical formulas to support controlled training and evaluation. However, strict automatic matching may penalize alternative but semantically valid formulations. Our manual analysis also suggests that some reference formulas may be underspecified or overly canonicalized, especially for requirements involving symmetric bounds, equivalent rewritings, or alternative signal-name conventions. Future work should consider multi-reference annotations and semantic-equivalence-aware evaluation protocols.

Third, tool-augmented reasoning introduces additional inference overhead. Compared with direct generation, explicit reasoning traces and tool calls increase decoding length and require external tool execution. Although the current tool set is deterministic and lightweight, latency may remain a practical concern for interactive use or large-scale batch processing. This motivates future work on adaptive tool invocation, more efficient reasoning policies, and selective use of computation only when it is necessary.

Fourth, the current experiments are conducted primarily with a Qwen3-4B backbone due to computational constraints. Larger open-source models, such as Qwen3-8B or Qwen3-30B-A3B, may further improve performance on complex multilingual requirements, deeply nested temporal structures, and longer tool-use chains. A systematic scaling study across model sizes, training budgets, and tool-use policies is left for future work.

Finally, although \textsc{ReasonSTL} substantially improves NL-to-STL generation, its current accuracy remains insufficient for unsupervised large-scale deployment in safety-critical CPS settings. Generated specifications may still contain subtle semantic errors, missing assumptions, overly restrictive or overly permissive constraints, or modeling choices that do not match the intended physical system. Therefore, \textsc{ReasonSTL} should currently be viewed as a human-in-the-loop drafting assistant rather than a replacement for expert specification design. In practical use, generated STL candidates should be reviewed by domain experts, checked against system-specific modeling assumptions, and validated through downstream formal methods before deployment.

\end{document}